\documentclass[a4paper]{report}
\usepackage[utf8]{inputenc}
\usepackage[T1]{fontenc}
\usepackage{RJournal}
\usepackage{amsmath,amssymb,array}
\usepackage{booktabs}

\usepackage{graphicx}%
\usepackage{multirow}%
\usepackage{amsmath,amssymb,amsfonts}%
\usepackage{amsthm}%
\usepackage{mathrsfs}%
\usepackage[figuresright]{rotating}%
\usepackage[title]{appendix}%
\usepackage{xcolor}%
\usepackage{textcomp}%
\usepackage{manyfoot}%
\usepackage{booktabs}%
\usepackage{algorithm}%
\usepackage{algorithmicx}%
\usepackage{algpseudocode}%
\usepackage{program}%
\usepackage{listings}%

\begin{document}

\sectionhead{DiscoVars}
\volume{XX}
\volnumber{22}
\year{2022}
\month{May}

\begin{article}
\title{DiscoVars: A New Data Analysis Perspective - Application in Variable Selection for Clustering}
\author{by Ayhan Demiriz}
\maketitle

\abstract{
We present a new data analysis perspective to determine variable importance regardless of the underlying learning task. Traditionally, variable selection is considered an important step in supervised learning for both classification and regression problems. The variable selection also becomes critical when costs associated with the data collection and storage are considerably high for cases like remote sensing. Therefore, we propose a new methodology to select important variables from the data by first creating dependency networks among all variables and then ranking them (i.e. nodes) by graph centrality measures. Selecting Top-$n$ variables according to preferred centrality measure will yield a strong candidate subset of variables for further learning tasks e.g. clustering. We present our tool as a Shiny app which is a user-friendly interface development environment. We also extend the user interface for two well-known unsupervised variable selection methods from literature for comparison reasons.
}

\section{Introduction}\label{sec1}
\setcounter{section}{1}

Applications of machine learning and related technologies are dependent on successful implementations of learning algorithms and strong data analytic skills of practitioners. Often complexity and black box nature of the algorithms disengage practitioners from the knowledge discovery process. Recently, high dimensional data and uninterpretable complex models are two major reasons that practitioners have started shying away from interacting with learning models and have preferred AutoML \cite{MartinezPlumed2021,Truong2019}. Incorporating domain knowledge and expertise into algorithmic models certainly allows practitioners to manage the learning process.

Feature (variable) selection is an important task and can be performed by employing a large array of methods in the literature \cite{Li2017,Guyon2003,Caruana2003,Forman2003}. Depending on learning task, supervised and unsupervised variable selection are possible. Considering that machine learning problems have been studied for several decades, many well known methods were originally developed under very limited computational resource constraints. Naturally, dimension reduction is one way of reducing problem complexity for the underlying methods. Feature extraction and feature selection are two alternative choices for dimension reduction \cite{Li2017}. The first alternative is typically used for projecting the full input space to a lower dimension. Feature extraction requires the usage of full input data in data preparation stage of the underlying learning method. Moreover, such methods may still require more computational resources and full data at later stages of the learning process. Feature selection, on the other hand, picks a subset of all variables presumably without any sacrifice in the performance of learning process. The purpose of feature selection is primarily to reduce the computational complexity and then to increase the performance of learning process which could be in supervised, unsupervised and semi-supervised fashions. Overfitting is also another reason to utilize feature selection to prevent very poor generalization in learning processes.

Categorization of the feature selection methods is done from various perspectives in practice \cite{Li2017,Guyon2003}. Supervision and selection strategy perspectives are the most common categorization of feature selection methods \cite{Li2017}. Depending on level of label data, supervised, unsupervised and semi-supervised approaches are utilized in feature selection. Specifically, label information is used in classification and regression problems. Subset of features are selected based on some performance measures such as accuracy and  $R^2$. In addition to level of supervision,  selection strategy is another categorization of feature selection methods. Selection could be implemented as a wrapper method which calls the learning model like a black box with different combinations of input data and searches for the best possible subset of variables. Since there are $2^d$ different input possibilities for $d$-dimensional data, the combinatoric nature of this approach may slow down the search. However, introducing some evaluation criteria may speed up the process. For example, traditional stepwise, backward and forward selection methods can be considered as wrapper methods in multivariate regression problems and these methods simply pick a feature based on its contribution to the overall $R^2$ value at each iteration. Usually one feature is added/subtracted from the regression model at each iterative search step. Obviously, supervision is also used. The second approach for selection strategy is filter methods. The main idea is to select a subset of variables by ranking them prior to running the learning algorithm. Potentially, filter methods can be used irrespective of underlying learning task i.e. supervised, unsupervised or semi-supervised. Ranking could be based on an intrinsic measure such as correlation and mutual information \cite{Chandrashekar2014}. Filter methods are composed of two steps: the first step ranks the variables according to some measure either univariate or multivariate (i.e. multiple features) way, the second step removes low ranked features \cite{Li2017}.  The third way of selection strategy is to embed feature selection into model learning \cite{Li2017}. This is somewhat a hybrid approach by combining best parts of previous two approaches.

We propose an interactive multivariate filter method that first creates a dependency network \cite{Heckerman2000} among all features then the nodes of this network are ranked based on centrality measure chosen by the user. For commonly used network centrality measures, interested user is referred to \cite{HANSEN202079}. Finally, Top-$n$ variables are selected by the user (modeler). Our approach enables modeler to discover candidate variable subset through an interactive method by utilizing dependency networks and graph centrality measures. Note that each node in dependency networks corresponds to a variable. Thus our method is named as DiscoVars (Discover Variables). Since supervised feature selection methods are established on concrete performance measures, we opt to implement our feature selection method for clustering analysis. Nevertheless, our approach can directly be used in classification problems as well. Novelty of our approach is to utilize graph centrality metrics to determine importance of variables on dependency networks constructed by efficient and proven variable selection methods such as stepwise, forward, and Lasso \cite{Tibs96}.

Throughout the paper,  $\cal{X}$ represents $m \times d$ dimensional dataset, $i$-th dimension of $\cal{X}$ is practically the random variable $X_i$, Top-$n$ variables are the selected $n$ features by the user (modeler), $\cal{S}$ represents $m \times n$ dimensionally reduced dataset, and $k$ is the number of clusters set for clustering algorithms, a dependency network is a directed graph $\mathcal{G}=(V,E)$ of  vertices $V$, edges $E$. The organization of the paper is as follows. Section \ref{sec2} introduces the methodology proposed in this paper. As an application domain, unsupervised feature selection is considered suitable for our methodology. Section \ref{sec3} reviews some related work in feature selection for clustering problems. Comparison of performances of clustering methods is very subjective by its nature. We report the detail implementation of our proposed approach DiscoVars in Section \ref{sec4}. For comparison reasons we also implemented two-well known feature selection methods on Shiny framework in Section \ref{sec5}. Section \ref{sec6} then concludes our paper.

\section{Methodology}\label{sec2}

Graphical models \cite{Koller2007} are considered as a popular tool for represent real world problems by combining uncertainty and logical structure i.e. conditional independence of variables. Bayesian and Markov Networks are the most common graphical models studied in the literature. Bayesian Networks are directed graphs. Markov Networks, on the other hand are undirected graphs. Graphical models primarily help on probabilistic inference. Thus, the goal is to derive a joint distribution $P$ over set of random variables $\mathscr{X}=\{X_1,\ldots,X_d\}$. Conditional independence is an important structural concept that simplifies underlying graphical models. According to Definition 2.1 of \cite{Koller2007}, conditional independence is defined as $ P(X=x,Y=y \mid  Z=z) = P(X=x \mid Z=z) P(Y=y \mid Z=z)$ for all $x$, $y$, and $z$ values. Assuming conditional independence  of variables $\{X_1,\ldots,X_d\}$ in a Bayesian Network (BN) $\cal{G}$, one can factorize joint probability distribution of BN as,

$$
P_B(X_1,\ldots,X_d)=\prod^d_{i=1} P(X_i \mid Pa_i),
$$
\noindent where $Pa_i$ are the parents of random variable $X_i$. Note that this rule is also called as chain rule of BN. A greedy algorithm that combines local and global structure search is proposed in \cite{Chickering1997} which utilizes conditional probability distributions. We implicitly assume that a random variable $X_i$ and its non-descendants are conditionally independent given $X_i$'s parents $Pa_i$. Markov Networks can also be constructed by factorizing over structural sub-graphs that are independent. Graphical models are computationally expensive to construct. In addition, it is very hard to interpret these models by untrained practitioners. Modelers prefer to interpret connections in these networks as relationships. However, relationships are casual and may not necessarily correspond to predictive ones.

Dependency networks (DNs) are considered as computationally efficient to construct \cite{Heckerman2000}. The primary intention to develop DNs was to visualize predictive relationships. Commercial version was available with MSSQL Server 2000. It is claimed in \cite{Heckerman2000} that it was very useful to discover dependency relationships as a graphical tool. By varying (increasing) the strength threshold of relationships, sparser networks can be drawn. The simplicity of a consistent DN comes from the fact that conditional distributions of variables $X_i$ can be found as:
$$
P(X_i \mid Pa_i)=P(X_i \mid \mathscr{X} \backslash X_i ),
$$
where $\mathscr{X} \backslash X_i$ means all the variables in $\mathscr{X}$ except $X_i$ and DN is consistent in a way that local probability distributions can be obtained from $P(\mathscr{X})$. It is shown in \cite{Heckerman2000} that consistent DNs are equivalent to Markov Networks. In other words, one can learn the structure of Markov Network from a consistent DN. This is an important result that makes DNs as practical graphical models for probabilistic inference, predicting preferences (collaborative filtering) \cite{Heckerman2000}, and sequential data inference (prediction) \cite{Carlson2008}. Constructing a DN may require using classification/regression models for estimating the local distributions \cite{Heckerman2000}. In commercial setting, probabilistic decision trees were implemented as a default learning method because of their simplicity and computational efficiency. Note that DNs are primarily used for supervised prediction problems. So, the underlying graph represents the significant relationships for the target (label) variable in consideration.

Our approach is primarily based on constructing a DN by fitting each variable $X_i$ with the remaining variables $\mathscr{X} \backslash X_i$ to represent all the significant relationships among all the variables. A statistical variable selection method $M$ should be used for determining significant variables for local distributions. Assume that the set $s$ represents the indices of significant variables for $X_i$ and the variables $X_s$ are parents $Pa_i$ of $X_i$. In other words, $X_i$ depends on variables $X_s$. All of these relationships form the DN \cal{G}. Note that we assume that all $X_i$ variables are continuous variables for this paper. Categorical data and classification methods can subsequently be incorporated into constructing DNs easily. So Stepwise, Forward, Akaike Information Criterion (AIC) and Lasso selection methods can be utilized by linear regression models (lm) in our approach at this time. The pseudo code of our methodology is given in Algorithm~\ref{algo1}.

\begin{algorithm}
\caption{DiscoVars: Construct DN $\cal{G}$}\label{algo1}
\begin{algorithmic}[1]
\State Given $\cal{X}$ and $M$
\For{$i \gets 1,\, d$}
\State $Fit~X_i \gets~lm(\mathscr{X} \backslash X_i,\, M)$
\State $s \gets \{significant~variables\}$
\State $E_{is} \gets 1 $
\EndFor

\State $c \gets Select~Centrality~Measure$
\State $Rank($ $\cal{G}$ $,\,c,\,$ $\mathscr{X})$
\State $Select~Top-n~variables$
\end{algorithmic}
\end{algorithm}
\bigskip

Interactive nature of our approach is based on choices of variable selection method $M$, centrality measure $c$ and $n$ for Top-$n$ important variables. The methodology given in Algorithm~\ref{algo1} is parallelizable due to the nature of \verb"for"  loop. However, one can argue that it is computationally expensive to construct $d$ regression models in the first place. Note that we can still apply filter methods to screen very high dimensional data ahead of constructing DNs if there is a consideration of computational resources. A stochastic search algorithm is proposed for constructing DNs on gene expression data for the purpose of supervised variable selection in \cite{Dobra2009}. Underlying graph structure among features can improve classification models for genome data \cite{SunChangLong2020}. By coupling graph structure with feature selection performs better than traditional feature selection methods \cite{SunChangLong2020}.

Centrality measures are used in our approach to rank nodes (variables) in DNs. It is a well established research topic to quantify structural importance of actors in a network \cite{Borgatti2006}. The centrality measures suitable for directed graphs such as betweenness, closeness, degree, eigenvector, and pagerank etc. can be used in our methodology. In common social network analysis problems, networks are composed of objects, people, or events. But features (variables) are the center of attention in our approach. In general, network science aims to study the structural relationships and importance in networks. Network topology determines the structural importance of nodes \cite{Roddenberry2020}. Centrality measures are usually computed based on full network topology. Some recent work enables inference of eigenvector rankings from data directly without inferring the full network topology \cite{Roddenberry2020,Roddenberry2021}. So one can argue that it is technically possible to rank variables without first constructing DNs, but this is not the scope of this paper. The full network topology is constructed first in this study to calculate centrality measures.

In \cite{Meinshausen2006}, local neighborhood structures are found by utilizing Lasso \cite{Tibs96}. The main idea in \cite{Meinshausen2006} is to estimate the graph structure of covariance matrix for high dimensional data. Similar to our approach local dependencies are found by Lasso. Note that that Lasso is an option in our implementation. Our approach basically differs from \cite{Meinshausen2006} once DN is constructed. In \cite{Meinshausen2006}, Lasso is used for constructing a sparse graph structure that all remaining variables in the graph become relevant for the learning task. Note that our approach is independent from learning task. Centrality measures are further used to filter most important variables (nodes) in the graph. There exist also distributional assumptions in \cite{Meinshausen2006}. The size of neighborhood in Lasso is very sensitive to the selection of regularization parameter $\lambda$. Generally, cross-validation is common way of setting the parameter. In extreme cases, very sparse and full connections may be found by Lasso. Some remedies are offered in \cite{Meinshausen2006} for Lasso to result in robust neighborhood formation i.e. connections in network. Therefore neighborhoods are formed in a more stable way compared to original Lasso formulation. Same situation may also be an issue in our approach. If network is extremely sparse or almost full in connections, graph centrality measures will not yield decisive rankings.

Structural properties of variable neighborhoods can also help computing Laplacian Scores \cite{He2005}. Variables can be ranked based on this score. This is another filter method that shares similar perspective with our work. Moreover, traveling salesman problem (TSP) is a universal test bed of ideas. In regular TSP, nodes (cities) have full connections. In other words, one can travel from any city to any other city directly. However, if sparse connections exist meaning that one can only move over physical road networks, we can rank cities based on underlying graph (network) structure. Thus, TSP can be solved by ranking cities first \cite{demirizTSP}. This is also very similar to our idea in this paper.

Our approach can be used as a filter method irrespective of underlying learning task. There are widely used and proven feature selection methods for supervised learning problems due to availability of robust performance measures. We think that applying our approach may make an impact on unsupervised feature selection. Therefore, $\cal{S}$ can easily be clustered by any clustering algorithm after applying Algorithm \ref{algo1}.

\section{Unsupervised feature selection}\label{sec3}

In practice, there exist some metrics to compare clustering results such as Davies-Bouldin Index (DBI) \cite{DavBould} and Adjusted Rand Index (ARI) \cite{HubArab85}. It is technically possible to devise a wrapper algorithm by incorporating some intelligence in search mechanism to run clustering methods to come up with best possible feature subset. These clustering indices are easy to interpret but they are not guaranteed to be useful for a robust search optimization. They are not only dependent on features selected but also number of partitions at the same time. Certainly, some heuristics methods can be deployed to find a suitable subset of features. However this is not within the scope of our paper.

An excellent review of feature selection methods for model-based clustering is given in \cite{Fop2018}. Technically, the aim of variable selection is to determine the set of relevant variables for clustering. Logically, the remaining variables are called irrelevant. Two major assumptions are local conditional independence assumption of relevant variables within clusters and global independence assumption of irrelevant and relevant variables. Model-based clustering assumes that each observation comes from a finite mixture of $G$ probability distributions. Obviously, each distribution represents a different group \cite{Raftery2006,Scrucca2018,ScruccaFop}. Bayesian Information Criterion (BIC) is commonly used in model (i.e. performance) comparison for clustering. BIC can be calculated as follows \cite{Raftery2006}.

$$
BIC=2*\log(\text{maximized likelihood})-(\text{no. of parameters})*\log(m).
$$

Technically, BIC values of inclusion and exclusion of variable $X_i$ can be compared and a decision is made regarding that variable \cite{Scrucca2018}.

A Lasso like approach is proposed in \cite{WittenTibs2010}. Technically, between cluster sum of squares for feature $X_i$ is optimized and regularization terms $L_1$ (Lasso) and $L_2$ are applied on weights of features. $L_2$ penalty term guarantees non-zero solution. $L_1$ penalty term forces for sparser solutions \cite{WittenTibs2010}.

\cite{WittenTibs2010} maximizes between cluster sum of squares. Alternatively, it is also possible to reduce within group variance \cite{Andrews2014}. VSCC method \cite{Andrews2014} selects iteratively those variables that have smaller within-group variance and are correlated. We believe that feature selection for model-based clustering is well-studied. Interested readers are referred to \cite{Fop2018}. In Section \ref{sec5}, two well-known methods are implemented in Shiny framework for comparison and reproducibility purposes.

\section{DiscoVars}\label{sec4}

\begin{figure}[ht]%
\centering
\includegraphics[width=0.9\textwidth]{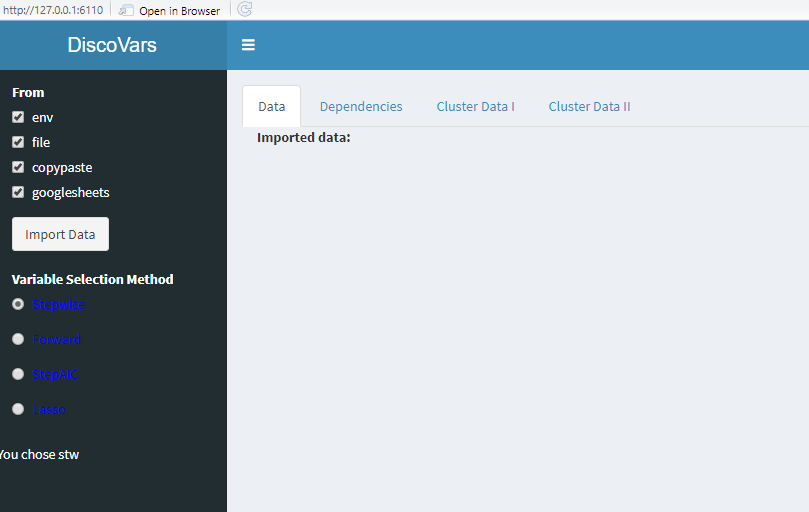}
\caption{DiscoVars Opening Screen}\label{Scr1}
\end{figure}

\begin{figure}%
\centering
\includegraphics[width=0.9\textwidth]{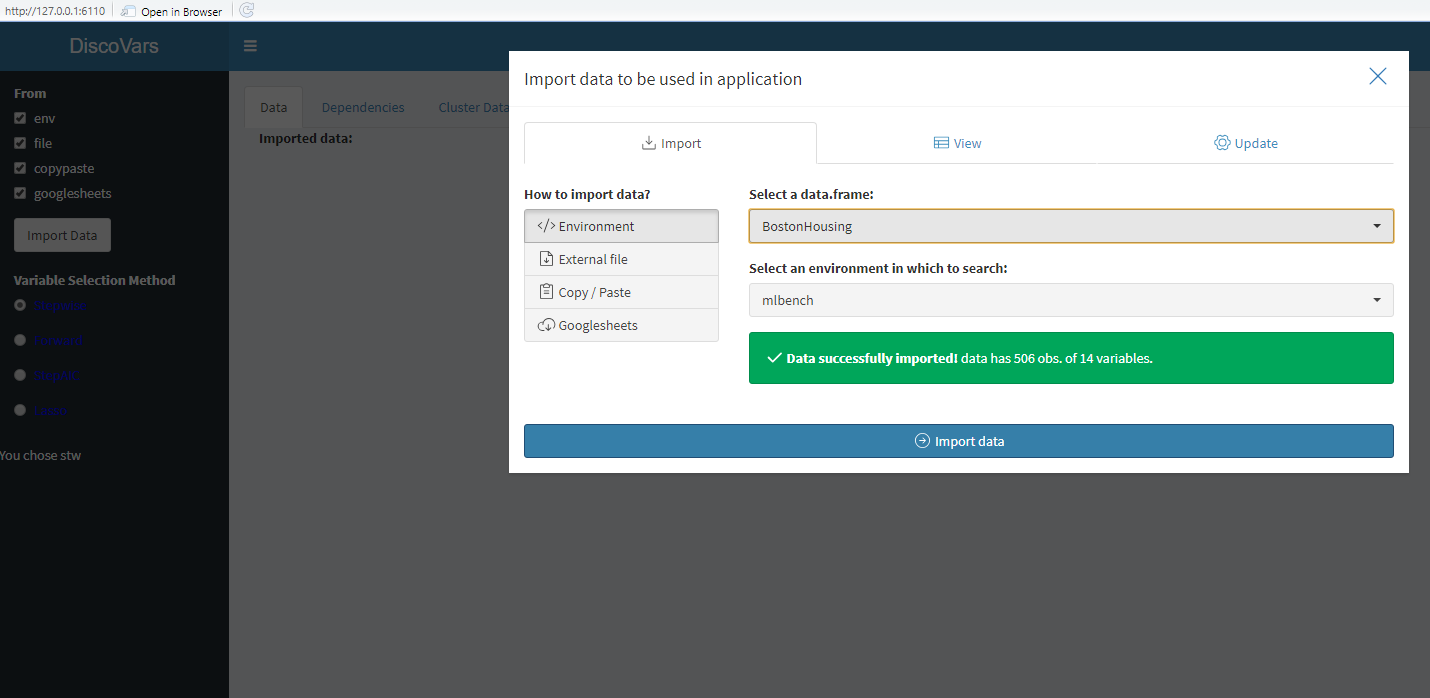}
\caption{Data Import Screen}\label{Scr3}
\end{figure}

Our approach is implemented by using Shiny \footnote{\url{https://shiny.rstudio.com/}} framework in R to design and develop interactive applications (Figure \ref{Scr1}). The current version of DiscoVars\footnote{Software runs on RStudio.} can be accessed at Github repository \footnote{\url{https://github.com/ademiriz/DiscoVars}}. As outlined in Algorithm \ref{algo1}, DiscoVars consists of two main steps:
\begin{itemize}
                                              \item Constructing DN
                                              \item Ranking and selecting Top-$n$ variables interactively according to choice of network centrality measure.
                                            \end{itemize}
DiscoVars can import data from various sources by utilizing \pkg{datamods}\footnote{\url{https://dreamrs.github.io/datamods/index.html}} R package (see Figure \ref{Scr3}). \pkg{datamods} is also a Shiny application. As seen in Figure \ref{Scr3}, user can import datasets available from other R packages such as \pkg{mlbench}. Notice that \verb"BostonHousing" Dataset from \pkg{mlbench} is used for illustration purposes.

\begin{figure}%
\centering
\includegraphics[width=0.9\textwidth]{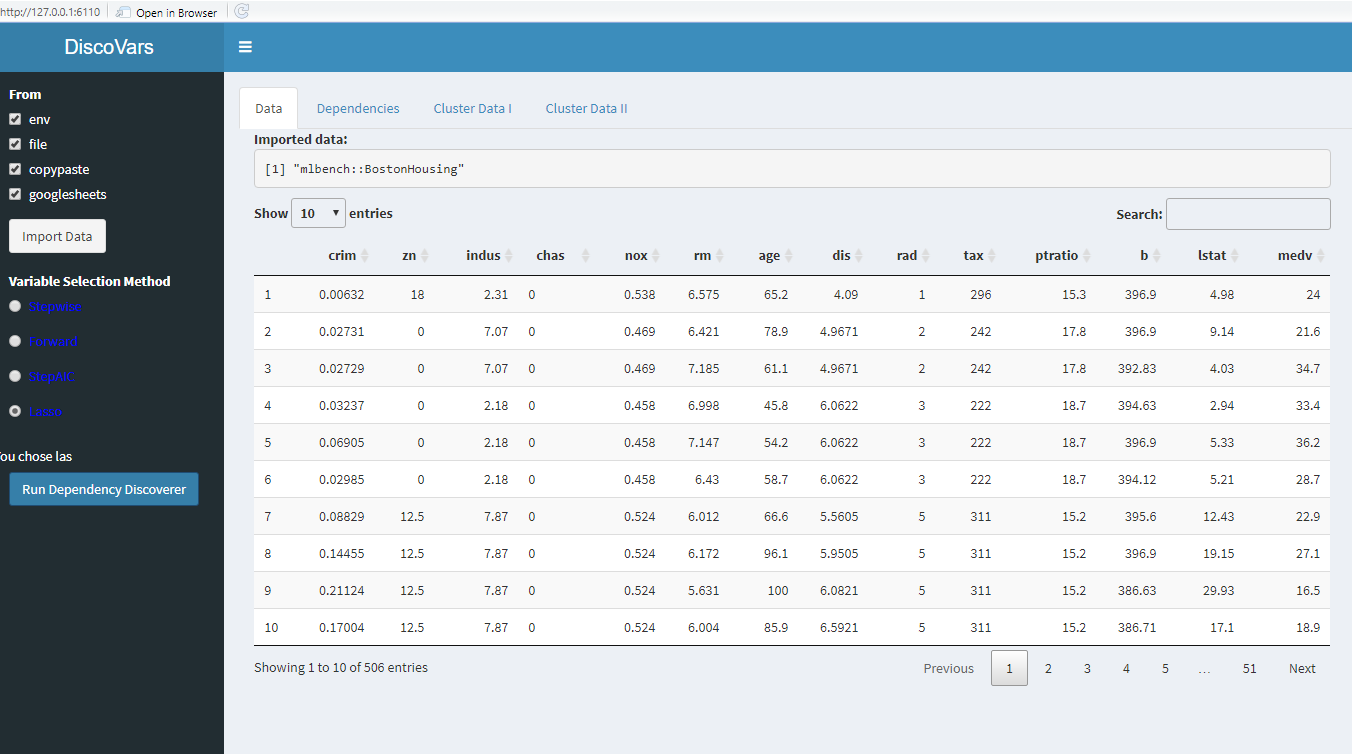}
\caption{Data Table}\label{Scr4}
\end{figure}

Note that only numeric variables are included in our implementation for the time being. Categorical variables can also be included via probabilistic decision tree models to construct DNs. Once the dataset selected by the user as in Figure \ref{Scr3}, user can filter data and/or exclude some unrelated variables (such as numerical ID variable) from dataset. Once dataset is imported, only numeric variables are used in DiscoVars. Imported dataset is presented to the user via \verb"dataTable" object (Figure \ref{Scr4}). At this moment, user can run dependency discoverer by choice of variable selection method $M$. Four well-known variable selection methods are available in DiscoVars. The default method is \verb"Stepwise"; \verb"Forward", \verb"stepAIC" and \verb"Lasso" are also presented to the user as alternatives. \verb"Stepwise" and \verb"Forward" methods are available from \verb"olsrr" package, \verb"stepAIC" is available from \pkg{MASS} package and \verb"Lasso" is available from \pkg{glmnet} package in R. For \verb"stepwise" selection, $p \leq 0.1$ entry and $p \geq 0.25$ exit parameters are set. For \verb"forward" selection, $p \leq 0.1$ entry parameter is set. Default parameter settings are used for \verb"stepAIC". With the default settings, \pkg{glmnet} runs Lasso with a varying number of $\lambda$ values. Therefore, a model selection is required. The parameter \verb"s" in \pkg{glmnet} is set to $16/m$ where $m$ is the number of rows. Cross-validation can be used in \pkg{glmnet} to pick the best model. This option is not use to avoid extended run times.

\begin{figure}%
\centering
\includegraphics[width=0.9\textwidth]{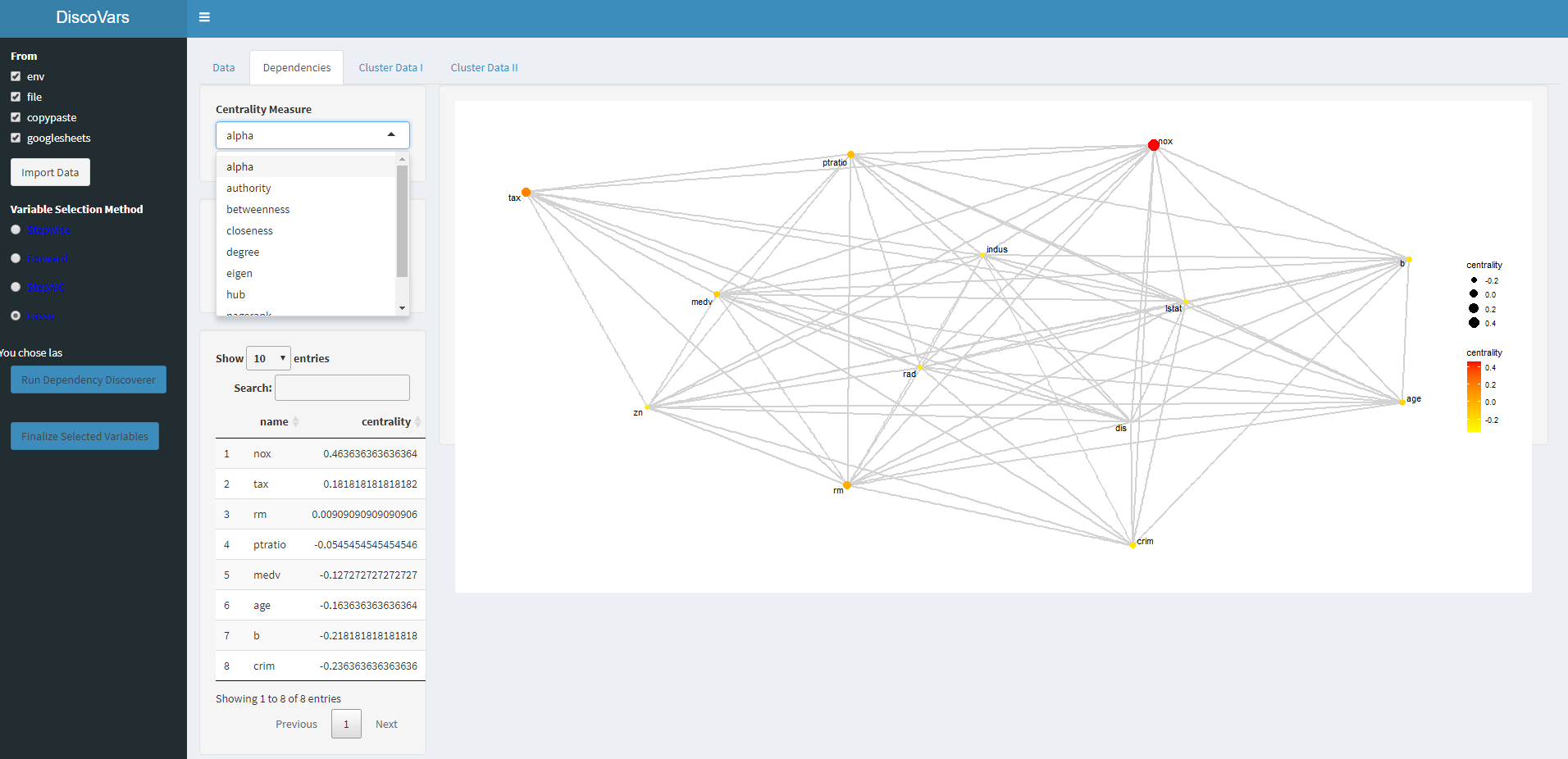}
\caption{Dependency Network and Setting Centrality Measure}\label{Scr5}
\end{figure}

\begin{figure}%
\centering
\includegraphics[width=0.9\textwidth]{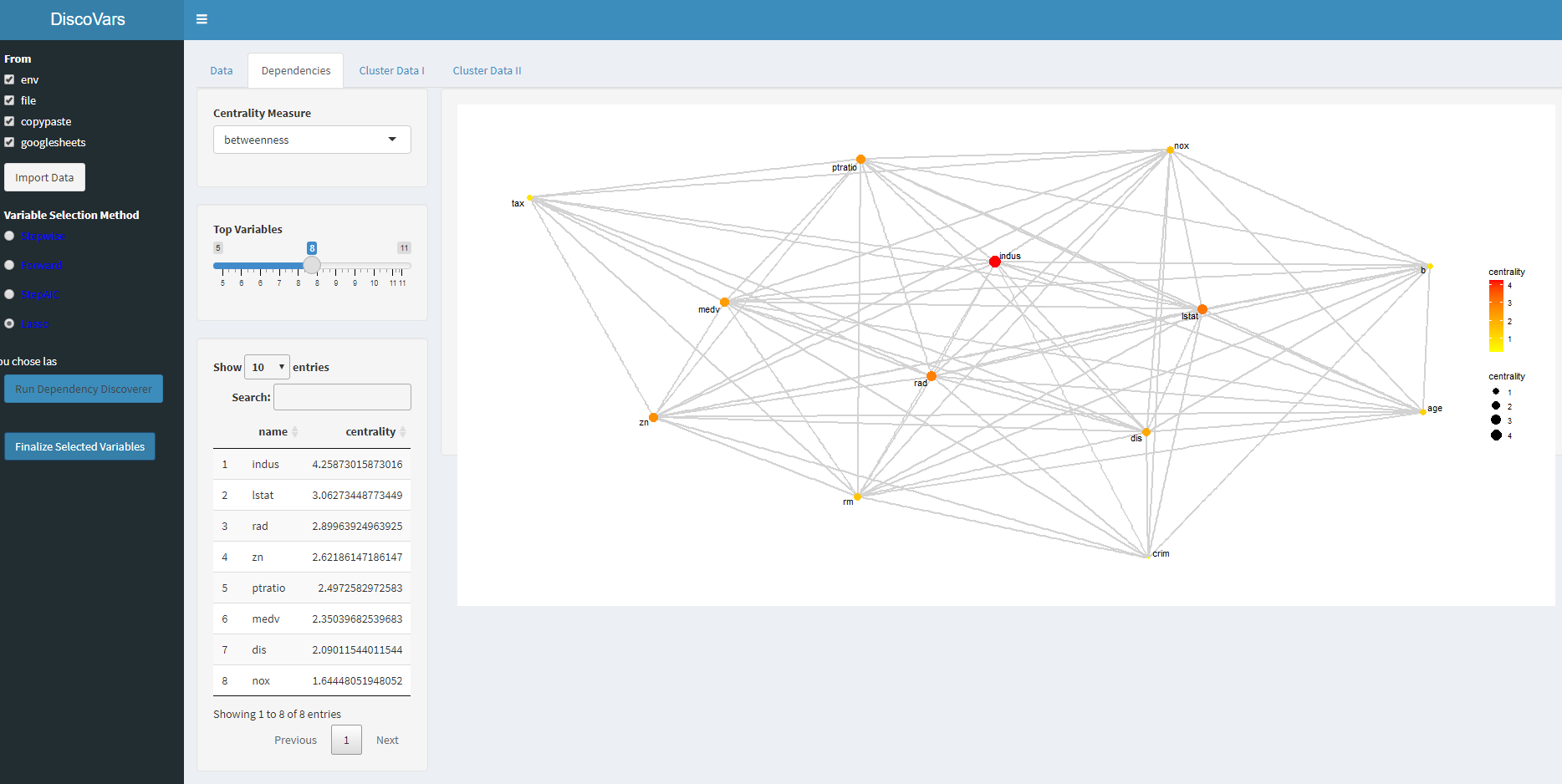}
\caption{Finalizing Top-$n$ Variables}\label{Scr6}
\end{figure}

DN can be constructed by running $d$ regression models as given in step 3 of Algorithm \ref{algo1}. Since this step is fully parallelizable, $d$ regression models are run by calling \verb"parSapply" function in \pkg{doParallel} R package. Parallelization obviously speeds up this step significantly. Figure \ref{Scr5} depicts the DN constructed. User is also able to redraw the network according to choice of centrality measures - \verb"alpha", \verb"authority", \verb"betweenness", \verb"closeness", \verb"degree", \verb"eigen", \verb"hub", \verb"pagerank", and  \verb"power". The default measure is \verb"alpha". Once user finalizes the centrality measure and $n$ for the variable selection, Top-$n$ variables are listed in a \verb"dataTable" object. $n$ is set by a \verb"sliderInput" object (see Figure \ref{Scr6}).

\begin{figure}%
\centering
\includegraphics[width=0.9\textwidth]{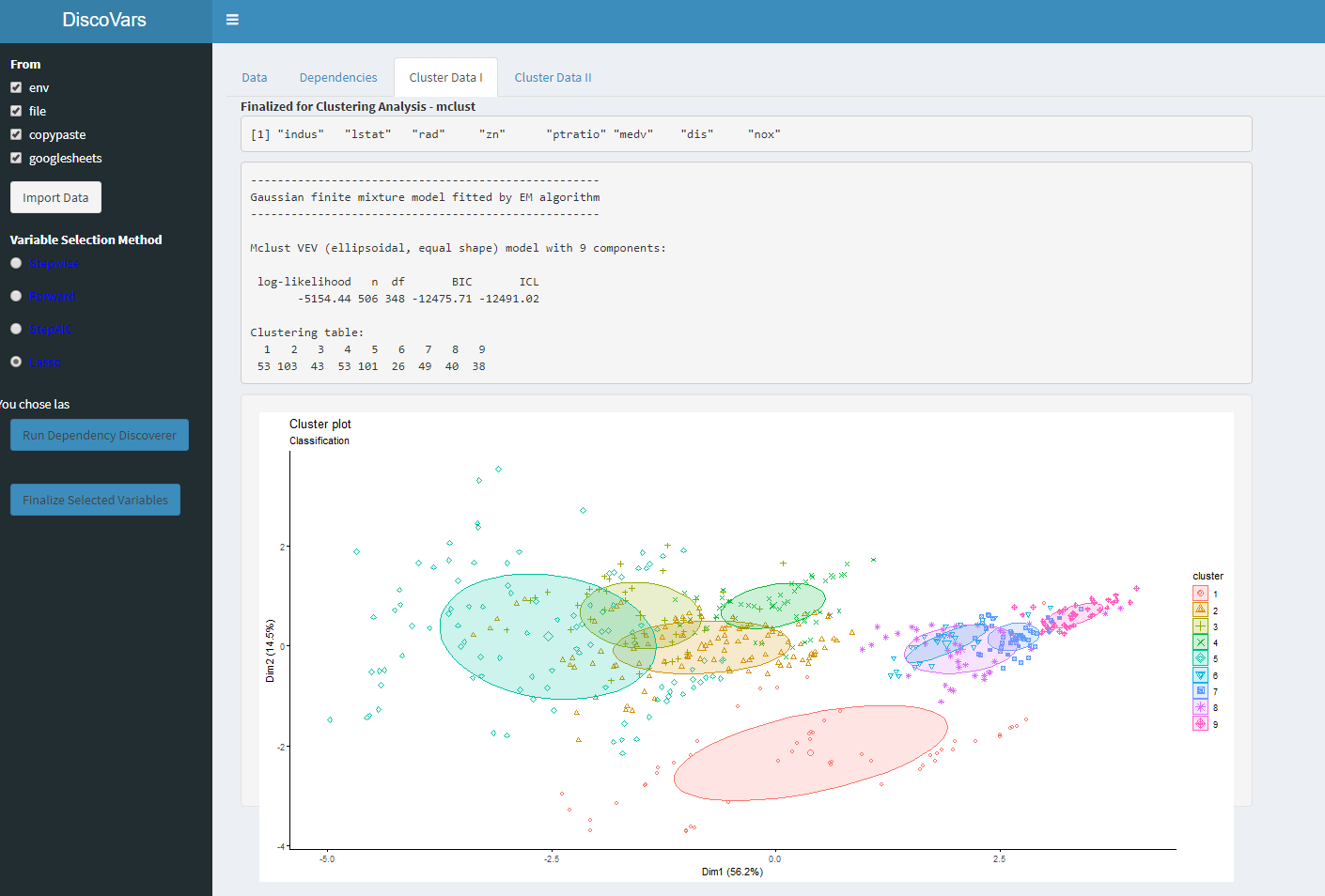}
\caption{Results of mclust Algorithm }\label{Scr7}
\end{figure}

\begin{figure}%
\centering
\includegraphics[width=0.9\textwidth]{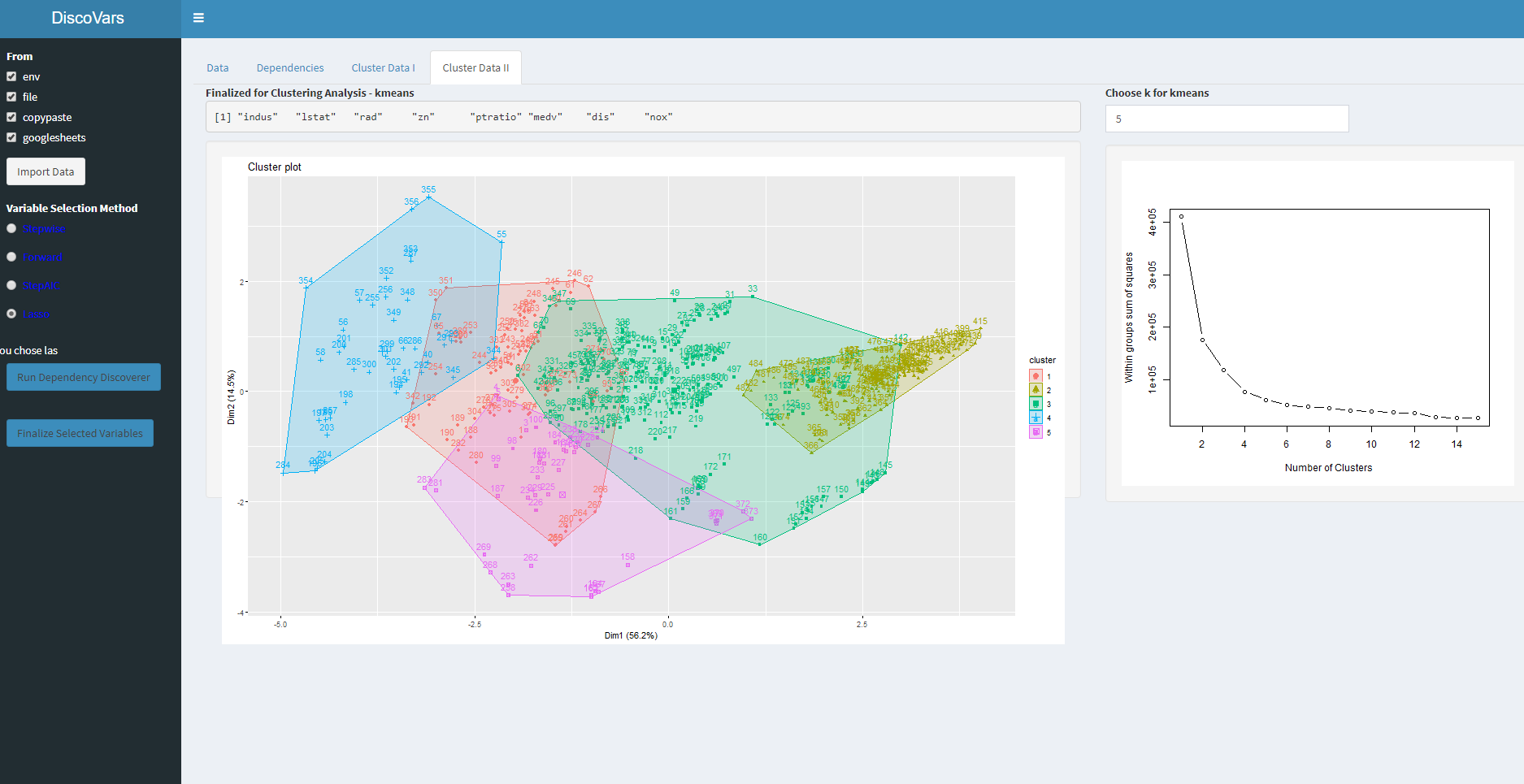}
\caption{Results of k-means Algorithm}\label{Scr8}
\end{figure}

After finalizing Top-$n$ variables, various clustering algorithms can be deployed to group data. \pkg{mclust} \cite{ScruccaFop} and \verb"k-means" algorithms are utilized in DiscoVars. Figures \ref{Scr7} and \ref{Scr8} depict outputs of \pkg{mclust} and \verb"k-means" respectively by using Top-$n$ variables. The default number of cluster parameter for \pkg{mclust} is 9 and it picks the best grouping with this initial condition. User can pick number of groups $k$ for \verb"k-means" algorithm (see Figure \ref{Scr8}). Elbow plot is also shown in \verb"k-means" output screen. Clustering results are plotted based on first two principal components in Figures \ref{Scr7} and \ref{Scr8}. \pkg{factoextra} package is used for this purpose. Notice that the whole process is interactive. Even if variable selection method, $M$, is changed DN will be reconstructed and Top-$n$ variables will be updated accordingly.

The most critical and time consuming part of DiscoVars is constructing DNs via $d$ regression models. DiscoVars can easily be used to filter important variables for numerical datasets. In order to report running times of DiscoVars for DN construction, we chose two different domains: anthropometric data and crypto currency market data. Anthropometry is the study of forms, measures and functional capacities of human body. We used two different datasets of anthropometric surveys on US military personnel \footnote{\url{https://www.openlab.psu.edu/data/}}. Measuring different parts of human body will cost proportionally to the number of measurements taken. The obvious question would be what measurements are most critical? The answer to this question requires expert domain knowledge. Note that this type of data is neither a classification nor a regression problem. Similar situation may arise in case of remote sensing data: which measurements are critical? We think that DiscoVars may yield reasonable results for this kind of cases without expert domain knowledge.

In the second set of domain, crypto currency market data are collected\footnote{\url{https://coinmarketcap.com/}}. Daily return rate of an asset, $r_t$ can be calculated as:

$$
r_t=\frac{v_t-v_{t-1}}{v_t},
$$
\noindent where $v_t$ and $v_{t-1}$ are daily closing values of an asset at dates $t$ and $t-1$ respectively.

We collected data and calculated daily return rates of two different sets of crypto currencies. In the first set, eight crypto coins' daily returns, first and second lags of these returns between 10-04-2017 and 09-24-2020 are calculated. In the second set, six crypto coins' daily returns, first and second lags of these returns between 08-10-2015 and 02-20-2018 are calculated. DiscoVars can principally discover major influencers among digital coins. Dimensions of both anthropometric and crypto currency market data are given in Table \ref{tab1}. These datasets are also provided with the software for reproducibility purpose\footnote{\url{https://github.com/ademiriz/DiscoVars}}.

\begin{table}
\begin{center}
\begin{minipage}{4in}
\caption{DN Construction Times for Various Variable Selection Methods on Representative Data}\label{tab1}%
\begin{tabular}{@{}lrrrrrr@{}}
\toprule
Dataset & $m$  & $d$ & Stepwise & Forward & stepAIC & Lasso\\
& & &  (sec.)  & (sec.) &  (sec.) & (sec.) \\
\midrule
Ansur Men\footnotemark[1]   & 1174   & 131 & 4881 & 6039 & 4751 & 5\\
Ansur Women\footnotemark[1]   & 2208  & 131 & 5780 & 7008 & 7805 & 6  \\
Ansur2 Men\footnotemark[2]   & 4082   & 93 & 2835 & 3391 & 3381 & 5 \\
Ansur2 Women\footnotemark[2]   & 1986  & 93  & 1972 & 2286 & 1001 & 5  \\
Coin I    & 1087   & 24  & 16 & 16 & 4 & 4  \\
Coin II    & 926  & 18  & 8 & 8 & 3 & 3  \\
\end{tabular}
\footnotetext[1]{\url{https://www.openlab.psu.edu/ansur/}}
\footnotetext[2]{\url{https://www.openlab.psu.edu/ansur2/}}
\end{minipage}
\end{center}
\end{table}

Table \ref{tab1} also summarizes running times of variable selection methods on anthropometric and crypto currency market data. DN construction times are reported in seconds. These experiments were run on a Windows 10 machine with fourth generation i7 processor, 16GB ram and R version 3.6.1. It is apparent that Lasso has a far better performance than the remaining methods. For small datasets, all methods can interactively be used in DiscoVars. For larger datasets, users are advised to utilize Lasso. Notice that parameters of variables selection methods are not tuned extensively in our implementation. Even stepwise regression can be sped up by lowering entry and exit $p$ values. Interested  users can easily modify provided code to reflect tuning variable selection methods.

By utilizing DiscoVars on \verb"Coin I" dataset with \verb"eigen" centrality measure, Top-5 variables are \verb"BNP_RTN", \verb"ETH_RTN_LG2", \verb"BNP_RTN_LG1", \verb"ETH_RTN_LG1", and \verb"BTC_RTN". Note that crypto currency datasets also include lagged returns.

\section{Implementation of clustvarsel and sparcl on Shiny}\label{sec5}

In order to have a comparison with our methodology, Shiny implementations of both \pkg{clustvarsel} \cite{Scrucca2018} and \verb"sparcl" \cite{WittenTibs2010} are also provided. Some information about these methods are already given in Section \ref{sec3}. We use a similar design framework for the implementations of \pkg{clustvarsel} and \verb"sparcl". User needs to import dataset first. Only numerical variables are included for analysis in both implementations. User is advised to remove univariate and ID variables again.

\begin{figure}%
\centering
\includegraphics[width=0.9\textwidth]{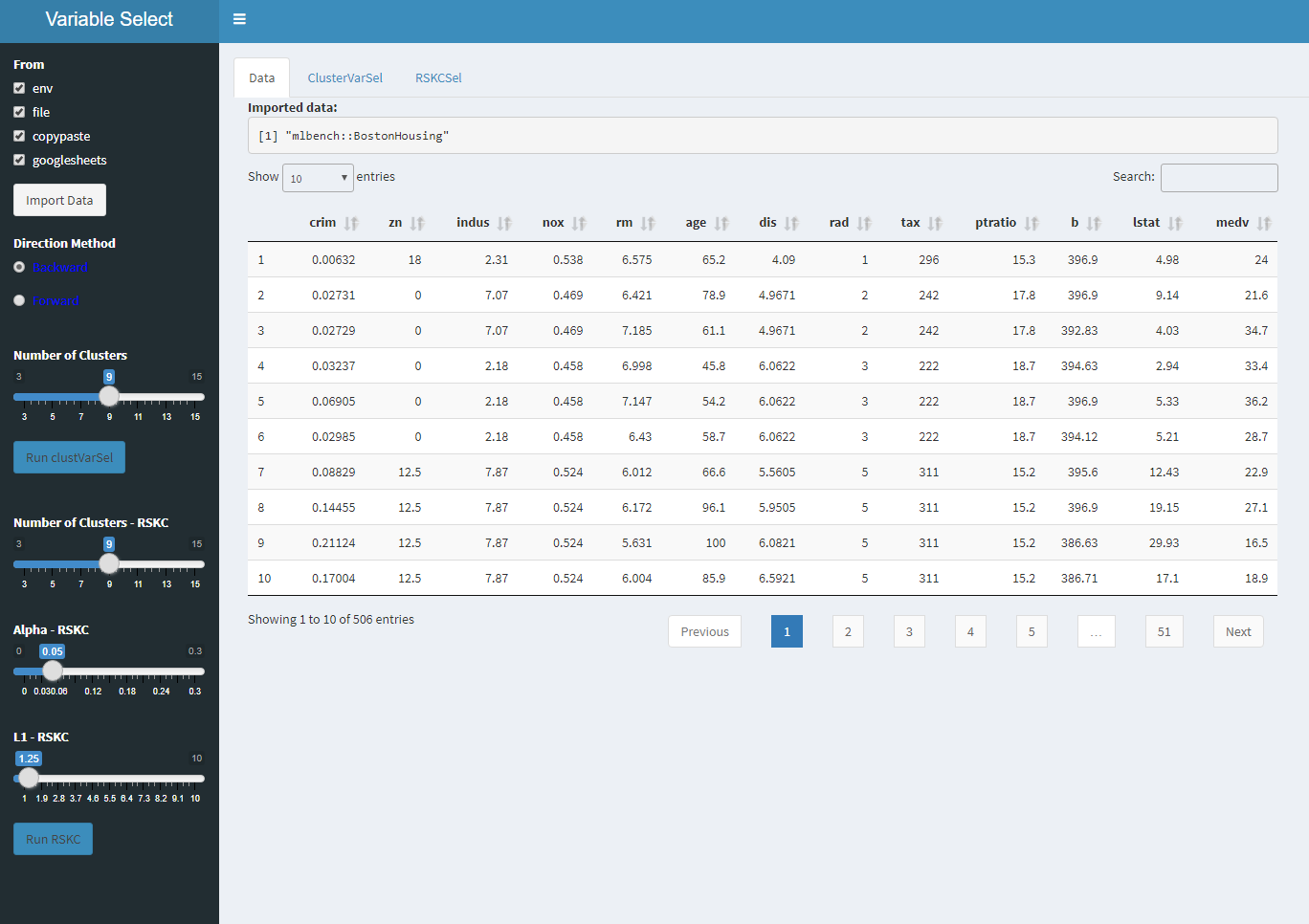}
\caption{Tool for Unsupervised Feature Selection Methods from Literature }\label{Scr9}
\end{figure}

\begin{figure}%
\centering
\includegraphics[width=0.9\textwidth]{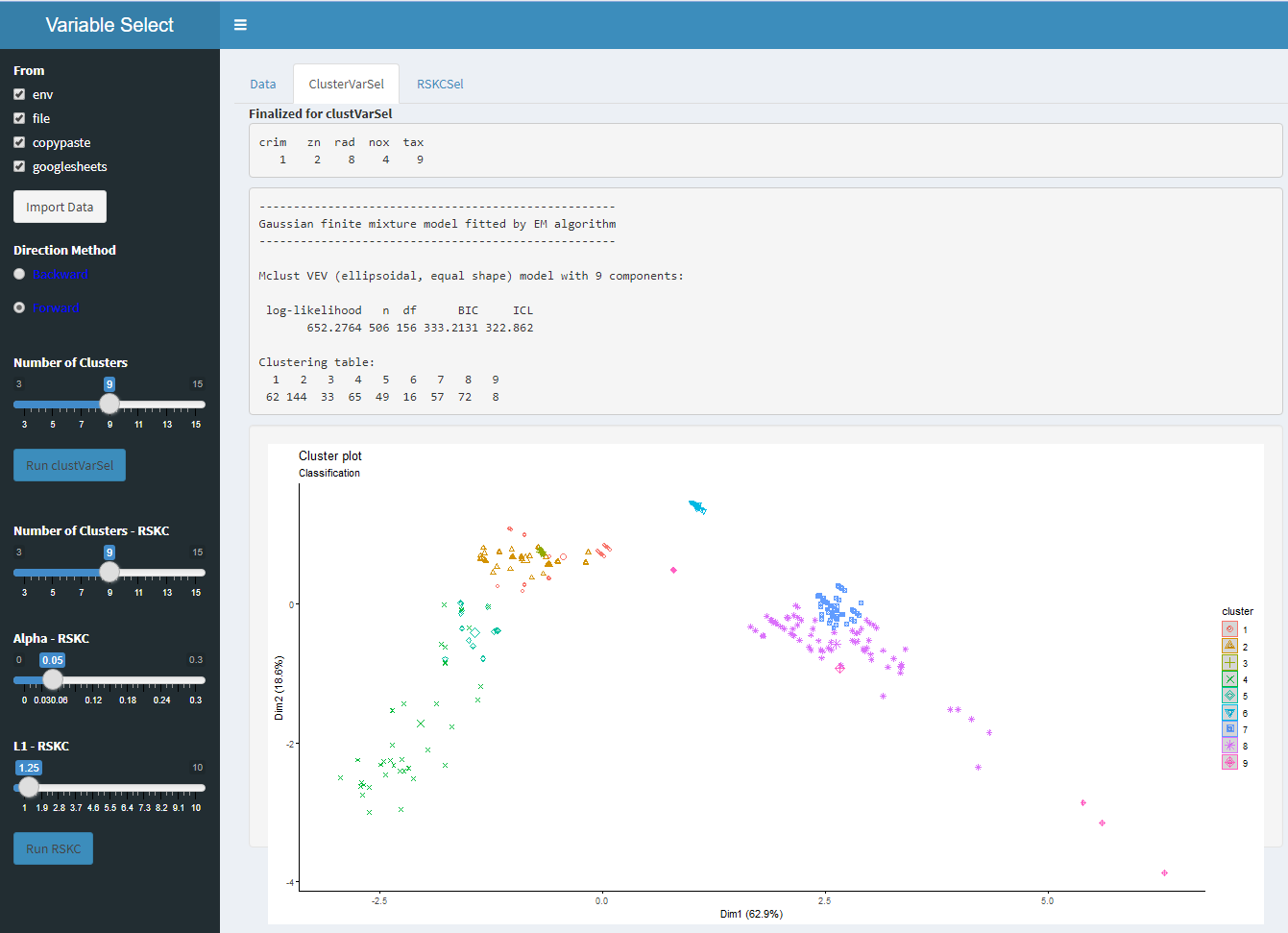}
\caption{clustvarsel Results}\label{Scr10}
\end{figure}

\pkg{clustvarsel} is based on \pkg{mclust} \cite{ScruccaFop} clustering method. Forward and backward directions i.e. variable inclusion and exclusion are available in this unsupervised feature selection method. The number of models parameter, $G$ can also be specified. The default value is 9. As seen in Figure \ref{Scr9}, user can pick search direction and number of clusters (models) parameters for our Shiny implementation.  Once \pkg{clustvarsel} is run, the output is shown to the user in Figure \ref{Scr10} like in \pkg{mclust} results (see Figure \ref{Scr7}). Note that variable selected by \pkg{clustvarsel} are shown in the text box at the top of Figure \ref{Scr10}. Since it is a wrapper method on top of \pkg{mclust}, \pkg{clustvarsel} may take longer times for high dimensional datasets. Inherently \pkg{mclust} is slower than \verb"k-means".

As a second method, \verb"sparcl" \cite{WittenTibs2010} is implemented in Shiny via \pkg{RSKC} Package \cite{Kondo2016} in R. \verb"sparcl" is available as stand alone R package, but \pkg{RSKC} is more preferable due to seamless integration with Shiny objects. As seen in Figure \ref{Scr9}, user is able to choose number of clusters, $\alpha$ and $L_1$ parameters for \pkg{RSKC} in our implementation interactively. Our initial experiments have indicated that results are highly sensitive to choice of parameters.

\begin{figure}%
\centering
\includegraphics[width=0.9\textwidth]{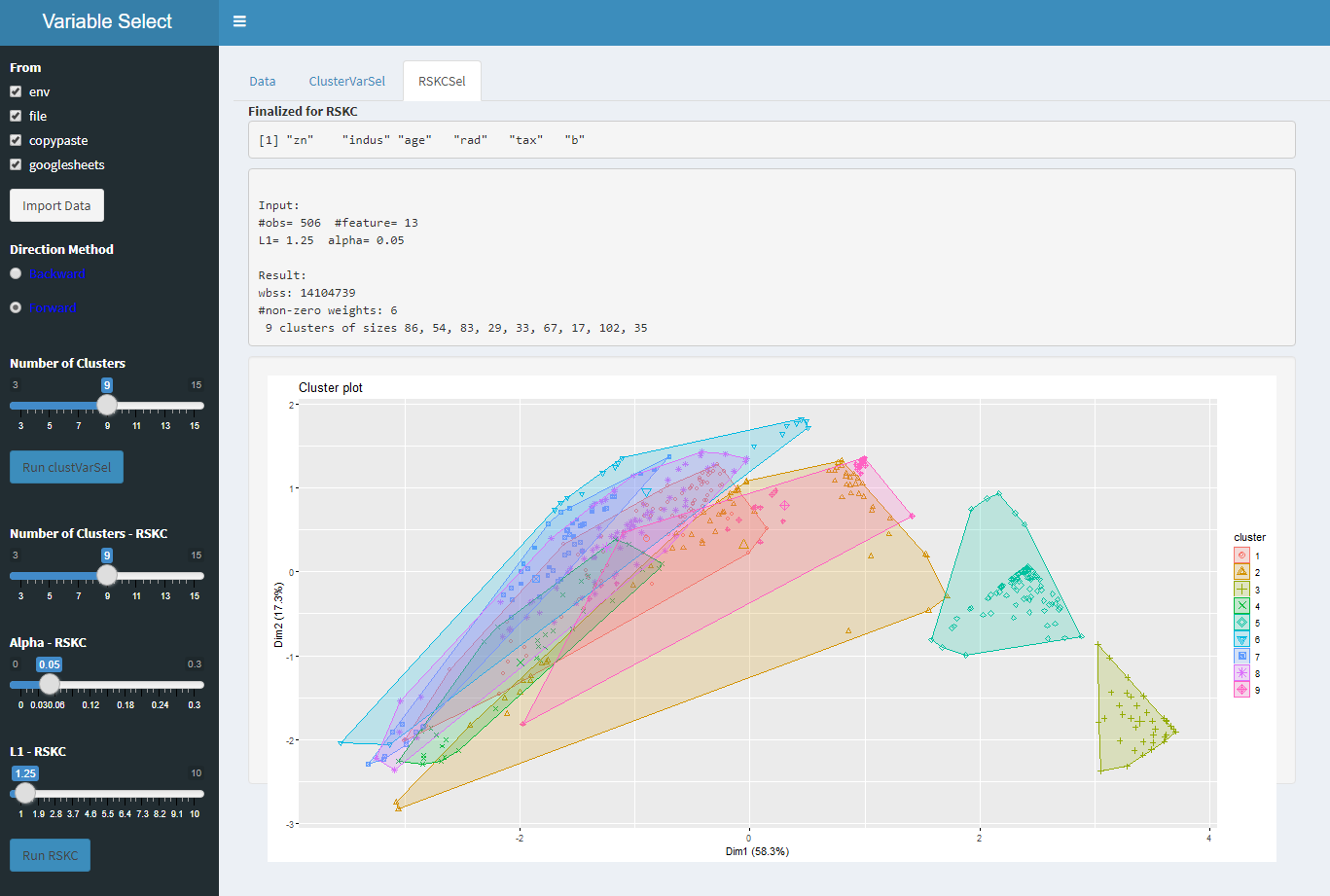}
\caption{RSKC Results }\label{Scr11}
\end{figure}

\begin{figure}%
\centering
\includegraphics[width=0.9\textwidth]{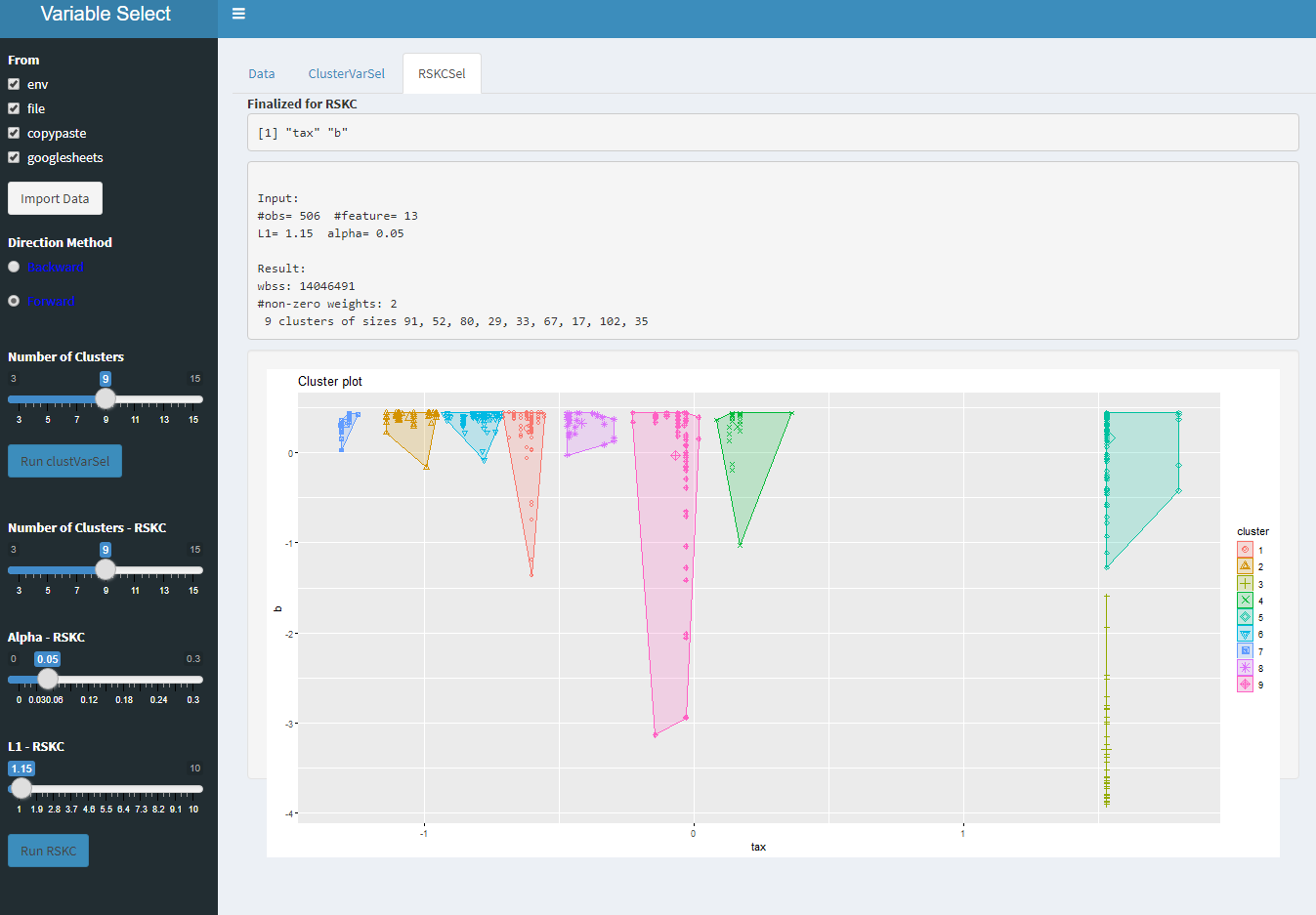}
\caption{Sparser RSKC Results}\label{Scr12}
\end{figure}

\begin{figure}%
\centering
\includegraphics[width=0.9\textwidth]{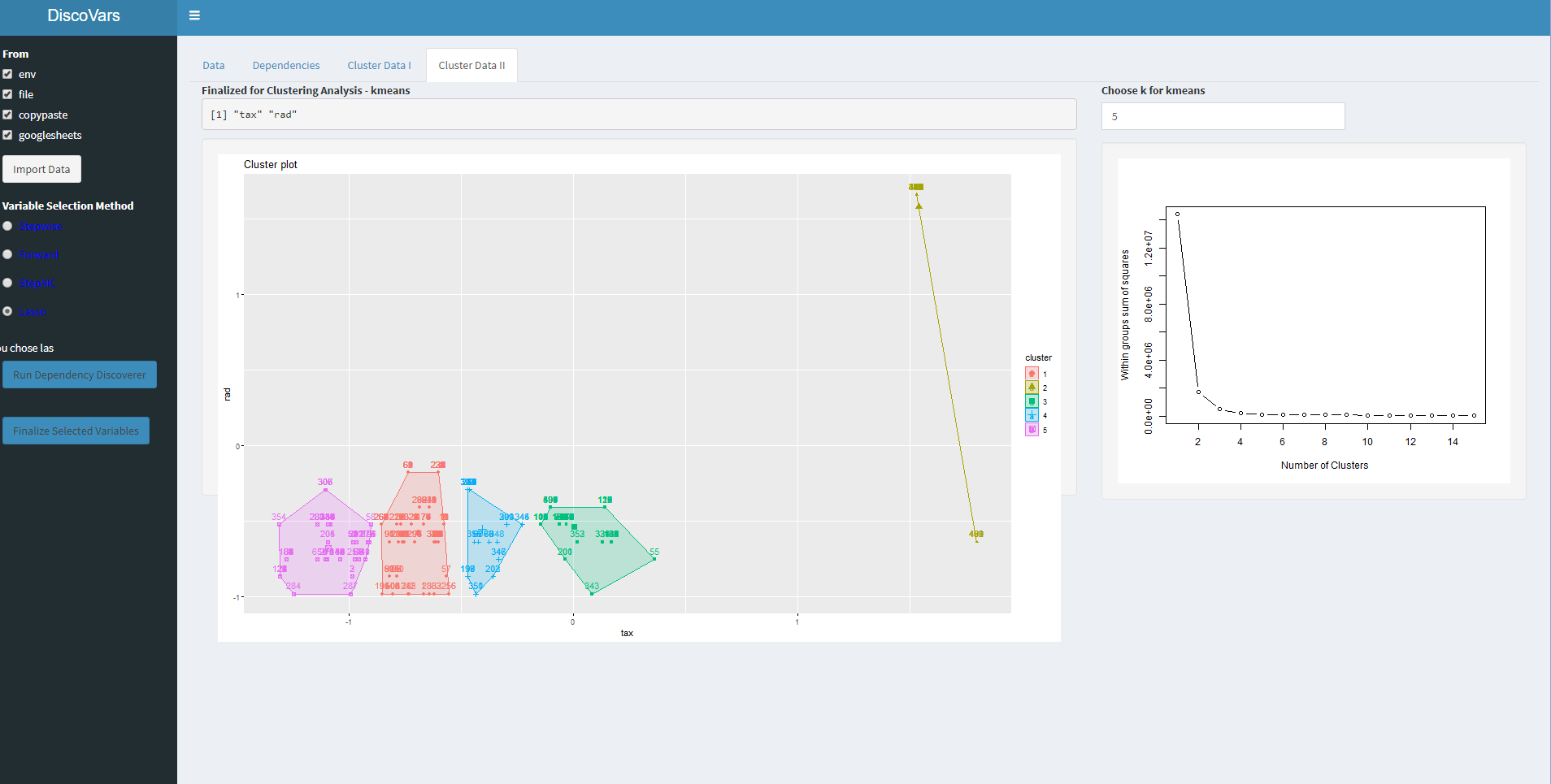}
\caption{Top-2 Variable Results from DiscoVars}\label{Scr13}
\end{figure}

Once \pkg{RSKC} is run, Figure \ref{Scr11} is shown to the user. The selected variables, standard output of \pkg{RSKC} and clustering plot (first two principal components with cluster labels) are presented. \verb"sparcl" formulation can be coupled with hierarchical clustering methods too, but \verb"k-means" is chosen as underlying clustering method. As user changes $L_1$ parameter, different variables may be chosen. For example, if we reduce $L_1$ to 1.15 from 1.25, we will get a sparser solution as in Figure \ref{Scr12}. If only variables \verb"tax" and \verb"b" are used for clustering BostonHousing dataset, we can get well-partitioned clusters. A similar result can be achieved with the \verb"authority" centrality measure by DiscoVars. If $n$ is set to 2, \verb"tax" and \verb"rad"  will be chosen (see Figure \ref{Scr13}).

The methods provided in this section are comparison reason. User can easily compare our approach, DiscoVars with methods from literature. User can import several datasets to try these methods interactively.

\section{Conclusions}\label{sec6}

In this paper, we presented a novel filtering method for learning algorithms. The steps of our general approach are
\begin{itemize}
  \item[i-] construction of a graphical model from the data set,
  \item[ii-] ranking of variables based on a centrality metric calculated by using the constructed graphical model, and
  \item[iii-] using the Top-$n$ variables for the learning algorithm.
\end{itemize}
 In the first step, we employ the well-known graphical model construction methods in the literature. Although there are studies concentrating on graphical model construction or centrality rankings (i.e. step (i) or (ii)), to the best of our knowledge, there is no paper combining both steps for the purpose of filtering for a learning algorithm. In our paper, we applied our new method to two real data sets (an anthropometric data set and a time series of crypto currency returns) and showed that our new method successfully returns satisfactory results in a reasonable computing time.

\bibliography{filename}

\address{Ayhan Demiriz\\
  Verikar Software\\
  Pendik, Istanbul, 34912\\
  Turkey\\
  (ORCiD 0000-0002-5731-3134)\\
  \email{ademiriz@gmail.com}}



\end{article}

\end{document}